\documentclass{article}
\usepackage{spconf,amsmath,graphicx}

\usepackage{graphicx}
\usepackage{amssymb,amsmath,bm}
\usepackage{textcomp}
\usepackage{times}
\usepackage{url}
\usepackage{latexsym}
\usepackage{color}
\usepackage{pgfplots}
\usepackage{subfigure,amssymb,amsmath,epsfig}
\usetikzlibrary{backgrounds,calc,positioning}
\usepackage{tikz,pgfplots,pgf}
\usetikzlibrary{matrix,shapes,arrows,positioning}

\usepackage{comment}
\usepackage{marvosym}
\usepackage{textcomp}

\def\x{{\mathbf x}}

\def\x{\textbf{x}}
\def\y{\textbf{y}}

\def\s{\textbf{s}}
\def\h{\textbf{h}}

\def\W{\textbf{W}}

\definecolor{greenForest}{RGB}{34, 139, 34}
\definecolor{purple}{RGB}{160, 32, 240}

\title{Parallel Long Short-Term Memory for Multi-stream Classification}

\name{Mohamed Bouaziz$^{1,2}$, Mohamed Morchid$^{1}$, Richard Dufour$^{1}$, Georges Linar\`es$^{1}$, Renato De Mori$^{1,3}$}
\address{
$^1$LIA - University of Avignon (France)\\ 
$^2$EDD - Paris (France)\\
$^3$McGill University - Montreal, Quebec (Canada)\\
}%

\begin{document}

  \maketitle

\begin{abstract}

Recently, machine learning methods have provided a broad spectrum of original and efficient algorithms based on Deep Neural Networks (DNN) to automatically predict an outcome with respect to a sequence of inputs. Recurrent hidden cells allow these DNN-based models to manage long-term dependencies such as Recurrent Neural Networks (RNN) and Long Short-Term Memory (LSTM). Nevertheless, these RNNs process a single input stream in one (LSTM) or two (Bidirectional LSTM) directions. But most of the information available nowadays is from multistreams or multimedia documents, and require RNNs to process these information synchronously during the training. This paper presents an original LSTM-based architecture, named Parallel LSTM (PLSTM), that carries out multiple parallel synchronized input sequences in order to predict a common output.
The proposed PLSTM method could be used for parallel sequence classification purposes. The PLSTM approach is evaluated on an automatic telecast genre sequences classification task and compared with different state-of-the-art architectures. 
Results show that the proposed PLSTM method outperforms the baseline n-gram models as well as the state-of-the-art LSTM approach.

\end{abstract}
\noindent{\bf Index Terms}: long short-term memory, sequence classification, stream structuring

  
\section{Introduction}
\label{sec:Introduction}



Recently, automatic sequence classification became an ubiquitous problem, having then encountered a high research interest 
~\cite{gers2001applying,severyn2015twitter,huang2016modeling}. This is due to the need to structure knowledge as a set of dependent localized information alongside with the new computer capabilities to efficiently process large amount of data. Among the recent methods employed to structure these sequences, the machine learning domain provides a set of high-level representations well adapted to automatic sequence classification based on Deep Neural Networks (DNN) such as Convolutional Neural Networks (CNN)~\cite{lecun1998gradient} or Recurrent Neural Networks (RNN)~\cite{elman1990finding}.


RNN architectures such as Long Short-Term Memory (LSTM)~\cite{hochreiter1997long} and Bidirectional LSTM (BLSTM)~\cite{graves2005framewise} have gained a particular attention in different domains and tasks including sentence~\cite{sundermeyer2012lstm} or successive images~\cite{vinyals2015show} processing. In speech recognition~\cite{graves2005bidirectional,fernandez2007application,wollmer2008abandoning}, these LSTM models exploit the contextual information whenever speech production or perception is influenced by emotion, strong accents, or background noise. The most effective use of RNNs for sequence classification is to combine the RNNs with Hidden Markov Models (HMMs) in a hybrid approach~\cite{bourlard2012connectionist,bengio1999markovian}. Nonetheless, RNNs or RNN-HMM could not be directly employed for sequence classification using multiple inputs from synchronous streams such as TV shows coming from different channels. Indeed, RNNs can only be trained to make a set of elements labeled in a single stream of input information.

In this paper, we introduce an original multistream neural network architecture, called Parallel LSTM (PLSTM), that simultaneously takes into account different synchronous streams in order to automatically classify this multistream sequence. To evaluate the effectiveness of the proposed PLSTM multistream neural network architecture, experiments are carried out on the LIA's Electronic Program Guide (EPG) dataset containing 3 years of TV programs from 4 different channels. The PLSTM performance 
is compared with the LSTM state-of-the-art approach as well as a classic n-gram approach considered as the baseline. Our PLSTM approach is an important step for sequence classification since it can be applied to any set of synchronous sequences.

Section~\ref{sec:rnn} proposes an overview of a couple of RNN architectures. Section~\ref{sec:pLstm} presents the proposed PLSTM. The experimental protocol and the discussion on the results are presented in Section~\ref{sec:experiments} and~\ref{s:Results} respectively. Finally, Section~\ref{sec:Conclusion} concludes this work and gives some interesting perspectives.



  
\section{Recurrent Neural Networks}
\label{sec:rnn}

This section introduces the state-of-the art concepts of two recurrent neural networks: LSTM and BLSTM.


\subsection{Long Short-Term Memory (LSTM)}
\label{sec:lstm}

Long Short-Term Memory (LSTM)~\cite{hochreiter1997long} networks are a special case of Recurrent Neural Networks (RNNs)~\cite{elman1990finding}. The goal of this architecture is to create an internal cell state of the network which allows it to exhibit dynamic temporal behavior. This internal state allows the RNN to process arbitrary sequences of inputs such as sequences of words~\cite{sundermeyer2012lstm} for language modeling, time series~\cite{gers2001applying}\dots 
The RNN takes as input a sequence $\x=(x_1,x_2,\dots,x_T)$ and computes the hidden sequence $\h=(h_1,h_2,\dots,h_T)$ as well as the output vector $\y=(y_1,y_2,\dots,y_T)$ by iterating from $t=1$ to $T$:
\begin{align}
h_t&=\mathcal{H}(\W_{xh}x_t+\W_{hh}h_{t-1}+b_h) \\
y_t&=\W_{hy}h_t+b_y
\end{align}

\begin{figure}
\begin{center}
\scalebox{0.5}{
\begin{tikzpicture}[
  font=\sffamily,
  every matrix/.style={ampersand replacement=\&,column sep=0.3cm,row sep=0.5cm},
  source/.style={draw,thick,rounded corners,fill=yellow!20,inner sep=.1cm},
  process/.style={draw,thick,circle,fill=blue!20},
  cel/.style={draw,thick,circle,scale=1},
  emptyCel/.style={draw,thick,circle,scale=1},
  sink/.style={source,fill=green!20},
  datastore/.style={draw,very thick,shape=datastore,inner sep=.3cm},
  dots/.style={gray},
  to/.style={->,>=stealth',shorten >=1pt,semithick,font=\sffamily\footnotesize},
  every node/.style={align=center}]

  \matrix{
    \& \node[source] (hisparcbox) {\LARGE{$x_t, h_{t-1}, b$}};  \\  
    
    \& \node[cel,minimum width=1cm] (it) {\LARGE{$i_t$}};  \& \& \& \node[cel,minimum width=1cm] (ot) {\LARGE{$o_t$}}; \& \\
    
    \node[cel] (ctanh) {\LARGE{$\alpha$}};  \& \node[cel, minimum width=0.1cm] (dit) {x}; \& \node[cel, minimum width=1.5cm,line width=3pt] (ct) {\LARGE{$c_t$}}; \& \node[cel] (otanh) {\LARGE{$\alpha$}}; \& \node[cel] (do) {x}; \&  \node[] (ht) {\LARGE{$h_t$}};\\
    
     \& \& \node[cel,minimum width=0.1cm] (df) {x};  \& \\
     \& \& \node[cel,minimum width=1cm] (ft) {\LARGE{$f_t$}};  \& \\
  };

\node [label={[shift={(-3.7,1.3)}] \large{Input gate} }] {};
\node [label={[shift={(1.2,1.3)}] \large{Output gate} }] {};
\node [label={[shift={(2.0,-3.4)}] \large{Forget gate} }] {};
\node [label={[shift={(-0.2,0.6)}] \large{Cell} }] {};

\draw[-triangle 45, line width=0.35mm] (hisparcbox) -- node[midway,above] {} node[midway,below] {} (it);
\draw[-triangle 45, line width=0.35mm] (hisparcbox) -- (-5.1,3.1) -- (-5.1,-0.1) -- node[midway,above] {} node[midway,below] {} (ctanh);
\draw[-triangle 45, line width=0.35mm] (hisparcbox) -- (2.95,3.1) -- node[midway,above] {} node[midway,below] {} (ot);
\draw[-triangle 45, line width=0.35mm] (hisparcbox) -- (-5.1,3.1) -- (-5.1,-3) -- node[midway,above] {} node[midway,below] {} (ft);
\draw[-triangle 45, line width=0.35mm] (it) -- node[midway,above] {} node[midway,below] {} (dit);
\draw[-triangle 45, line width=0.35mm] (ctanh) -- node[midway,above] {} node[midway,below] {} (dit);
\draw[-triangle 45, line width=0.35mm] (dit) -- node[midway,above] {} node[midway,below] {} (ct);
\draw[-triangle 45, line width=0.35mm] (ct) -- node[midway,above] {} node[midway,below] {} (otanh);
\draw[-triangle 45,dashed, line width=0.35mm] (ct) -- node[midway,above] {} node[midway,below] {} (it);
\draw[-triangle 45, line width=0.35mm] (ct) -- node[midway,above] {} node[midway,below] {} (ot);
%
\draw[-triangle 45, line width=0.35mm] (otanh) -- node[midway,above] {} node[midway,below] {} (do);
\draw[-triangle 45, line width=0.35mm] (ot) -- node[midway,above] {} node[midway,below] {} (do);
\draw[-triangle 45, line width=0.35mm] (ft) -- node[midway,above] {} node[midway,below] {} (df);
\draw[-triangle 45, line width=0.35mm] (do) -- node[midway,above] {} node[midway,below] {} (ht);
\draw[-triangle 45, line width=0.35mm]
    (ct) edge[bend left, dashed]  (df)
    (df) edge[bend left] (ct);
\draw[-triangle 45]
    (ct) edge[bend right=45,dashed] (ft);
\end{tikzpicture}
}
 \caption{Long Short-Term Memory (LSTM) cell. Dashed arrows correspond to connections with time-lag ($t-1$). $\alpha$ input/output activation function is usually $\tanh$.}
 \label{fig:lstm}
\end{center}
\end{figure}
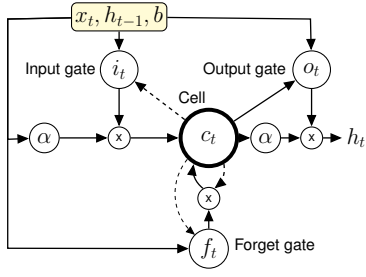

where $T$ is the total number of sequences; $\W_{xh}$ are the weight matrices between the input layers $\x$ and $\h$ and so on; $b$ is a bias vector, and $\mathcal{H}$ is the composite function.~\cite{hochreiter1997long} shows that LSTM networks outperform RNNs for finding long range context and dependencies. The LSTM composite function $\mathcal{H}$ forming the LSTM cell with peephole connections~\cite{gers2003learning} is presented in Figure~\ref{fig:lstm} and defined as:
\begin{align}
i_t&=\sigma(\W_{xi}x_t+\W_{hi}h_{t-1}+\W_{ci}c_{t-1}+b_i) \\
f_t&=\sigma(\W_{xf}x_t+\W_{hf}h_{t-1}+\W_{cf}c_{t-1}+b_f) \\
c_t&=f_tc_{t-1}+i_t\tanh(\W_{xc}x_t+\W_{hc}h_{t-1}+b_c) \\
o_t&=\sigma(\W_{xo}x_t+\W_{ho}h_{t-1}+\W_{co}c_{t}+b_o) \\
h_t&=o_t\tanh(c_t)
\label{equ:lstmFormula}
\end{align}

where $i$, $f$ and $o$, are respectively the input, forget and output gates, and $c$ the cell activation vector with the same size than the hidden vector $h$. The weight matrices $\W$ from cell $c$ to gates $i$, $f$ and $o$, are diagonal, and thus, an element $e$ in each gate vector receives only the element $e$ from the cell vector. Finally, $\sigma$ is the logistic sigmoid function.


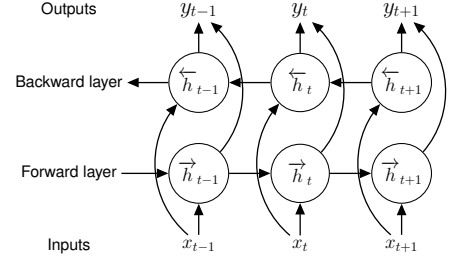
\begin{figure}
\begin{center}
\scalebox{0.5}{
\begin{tikzpicture}[
  font=\sffamily,
  every matrix/.style={ampersand replacement=\&,column sep=1.1cm,row sep=0.8cm},
  source/.style={draw,thick,rounded corners,fill=yellow!20,inner sep=.1cm},
  process/.style={draw,thick,circle,fill=blue!20},
  cel/.style={draw,thick,circle,scale=1},
  emptyCel/.style={draw,thick,circle,scale=1},
  sink/.style={source,fill=green!20},
  datastore/.style={draw,very thick,shape=datastore,inner sep=.3cm},
  dots/.style={gray},
  to/.style={->,>=stealth',shorten >=1pt,semithick,font=\sffamily\footnotesize},
  every node/.style={align=center}]

  \matrix{
\node[] (outputs) 
{\large{Outputs}};  \& \node[] (ytm1) {\LARGE{$y_{t-1}$}};  \& \node[] (yt) {\LARGE{$y_{t}$}}; \& \node[] (ytp1) {\LARGE{$y_{t+1}$}}; \\
\node[] (backward) 
{\large{Backward layer}};  \& \node[cel,minimum width=1cm] (htm1l) {\Large{$\overleftarrow{h}_{t-1}$}};  \& \node[cel,minimum width=1.55cm] (htl) {\Large{$\overleftarrow{h}_{t}$}}; \& \node[cel,minimum width=1.55cm] (htp1l) {\Large{$\overleftarrow{h}_{t+1}$}}; \\
\node[] (forward) 
{\large{Forward layer}};   \& \node[cel,minimum width=1.55cm] (htm1r) {\Large{$\overrightarrow{h}_{t-1}$}};  \& \node[cel,minimum width=1.55cm] (htr) {\Large{$\overrightarrow{h}_{t}$}}; \& \node[cel,minimum width=1.55cm] (htp1r) {\Large{$\overrightarrow{h}_{t+1}$}}; \\
\node[] (intputs) {\large{Inputs}};    \& \node[] (xtm1) {\Large{$x_{t-1}$}};  \& \node[] (xt) {\Large{$x_{t}$}}; \& \node[] (xtp1) {\Large{$x_{t+1}$}}; \\
};


%
\draw[-triangle 45, line width=0.35mm] (xtm1) -- node[midway,above] {} node[midway,below] {} (htm1r);
\draw[-triangle 45, line width=0.35mm] (xt) -- node[midway,above] {} node[midway,below] {} (htr);
\draw[-triangle 45, line width=0.35mm] (xtp1) -- node[midway,above] {} node[midway,below] {} (htp1r);
\draw[-triangle 45, line width=0.35mm] (htm1l) -- node[midway,above] {} node[midway,below] {} (ytm1);
\draw[-triangle 45, line width=0.35mm] (htl) -- node[midway,above] {} node[midway,below] {} (yt);
\draw[-triangle 45, line width=0.35mm] (htp1l) -- node[midway,above] {} node[midway,below] {} (ytp1);
\draw[-triangle 45, line width=0.35mm] (htm1l) -- node[midway,above] {} node[midway,below] {} (backward);
\draw[-triangle 45, line width=0.35mm] (htl) -- node[midway,above] {} node[midway,below] {} (htm1l);
\draw[-triangle 45, line width=0.35mm] (htp1l) -- node[midway,above] {} node[midway,below] {} (htl);
\draw[-triangle 45, line width=0.35mm] (forward) -- node[midway,above] {} node[midway,below] {} (htm1r);
\draw[-triangle 45, line width=0.35mm] (htm1r) -- node[midway,above] {} node[midway,below] {} (htr);
\draw[-triangle 45, line width=0.35mm] (htr) -- node[midway,above] {} node[midway,below] {} (htp1r);
%
%
%
%
\draw[-triangle 45, line width=0.35mm] (xtm1) edge[bend left=45] (htm1l) (xt) edge[bend left=45]  (htl) (xtp1) edge[bend left=45]  (htp1l)
    (htm1r) edge[bend right=45]  (ytm1)
    (htr) edge[bend right=45]  (yt)
    (htp1r) edge[bend right=45]  (ytp1);

\end{tikzpicture}
}
 \caption{Bidirectional Recurrent Neural Network (BRNN).}
 \label{fig:bdlstm}
\end{center}
\end{figure}

\subsection{Bidirectional Long Short-Term Memory (BLSTM)}
\label{sec:bdlstm}

LSTM networks use only the previous context to predict the next segment for a given sequence.
Bidirectional RNN (BRNN)~\cite{schuster1997bidirectional}, presented in Figure~\ref{fig:bdlstm}, can process both directions with two separate hidden layers (one for each direction). 
This type of RNN feeds to a same output layer fed forwarded inputs through the two hidden layers. 
Therefore, the BRNN computes both {\it forward} hidden sequence $\overrightarrow{\h}$ and {\it backward} sequence $\overleftarrow{\h}$ as well as the output vector $\y$, by iterating $\overrightarrow{\h}$ from $t=1$ to $T$, and $\overleftarrow{\h}$ from $t=T$ to $1$:
\begin{align}
\overrightarrow{h}_t&=\mathcal{H}(\W_{x\overrightarrow{h}}x_t+\W_{\overrightarrow{h}\overrightarrow{h}}\overrightarrow{h}_{t-1}+b_{\overrightarrow{h}}) \\
\overleftarrow{h}_t&=\mathcal{H}(\W_{x\overleftarrow{h}}x_t+\W_{\overleftarrow{h}\overleftarrow{h}}\overleftarrow{h}_{t+1}+b_{\overleftarrow{h}}) \\
y_t&=\W_{\overrightarrow{h}y}\overrightarrow{h}_t+\W_{\overleftarrow{h}y}\overleftarrow{h}_t+b_y
\end{align}

 By replacing the BRNN cells with LSTM cells, the Bidirectionnal LSTM (BLSTM)~\cite{graves2005framewise} is obtained. The BLSTM allows to exhibit long range context dependencies and takes advantage from the two directions structure. The output vector $\y$ is processed by evaluating simultaneously the two directions hidden sequences by computing the composite function $\mathcal{H}$ in the forward  ($\overrightarrow{\h}$) and backward ($\overleftarrow{\h}$) directions.

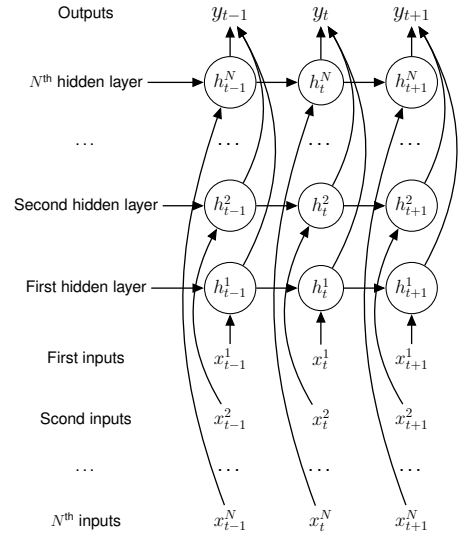
\begin{figure}
\begin{center}
\scalebox{0.5}{
\begin{tikzpicture}[
  font=\sffamily,
  every matrix/.style={ampersand replacement=\&,column sep=1.1cm,row sep=0.8cm},
  source/.style={draw,thick,rounded corners,fill=yellow!20,inner sep=.1cm},
  process/.style={draw,thick,circle,fill=blue!20},
  cel/.style={draw,thick,circle,scale=1},
  emptyCel/.style={draw,thick,circle,scale=1},
  sink/.style={source,fill=green!20},
  datastore/.style={draw,very thick,shape=datastore,inner sep=.3cm},
  dots/.style={gray},
  to/.style={->,>=stealth',shorten >=1pt,semithick,font=\sffamily\footnotesize},
  every node/.style={align=center}]

  \matrix{
  
\node[] (outputs) {\large{Outputs}};  \& \node[] (ytm1) {\LARGE{$y_{t-1}$}};  \& \node[] (yt) {\LARGE{$y_{t}$}}; \& \node[] (ytp1) {\LARGE{$y_{t+1}$}}; \\

\node[] (hN) {\large{$N^{\text{th}}$ hidden layer}};  \& \node[cel,minimum width=1.2cm] (htm1N) {\Large{$h_{t-1}^N$}};  \& \node[cel,minimum width=1.2cm] (htN) {\Large{$h_{t}^N$}}; \& \node[cel,minimum width=1.2cm] (htp1N) {\Large{$h_{t+1}^N$}}; \\
% ...
\node[] (h1) {\large{\dots}};  \& \node[] (htm11) {\Large{\dots}};  \& \node[] (ht1) {\Large{\dots}}; \& \node[] (htp11) {\Large{\dots}}; \\
\node[] (h2) {\large{Second hidden layer}};  \& \node[cel,minimum width=1.2cm] (htm12) {\Large{$h_{t-1}^2$}};  \& \node[cel,minimum width=1.2cm] (ht2) {\Large{$h_{t}^2$}}; \& \node[cel,minimum width=1.2cm] (htp12) {\Large{$h_{t+1}^2$}}; \\
\node[] (h1) {\large{First hidden layer}};  \& \node[cel,minimum width=1.2cm] (htm11) {\Large{$h_{t-1}^1$}};  \& \node[cel,minimum width=1.2cm] (ht1) {\Large{$h_{t}^1$}}; \& \node[cel,minimum width=1.2cm] (htp11) {\Large{$h_{t+1}^1$}}; \\
\node[] (intputs) {\large{First inputs}};    \& \node[] (xtm11) {\Large{$x_{t-1}^1$}};  \& \node[] (xt1) {\Large{$x_{t}^1$}}; \& \node[] (xtp11) {\Large{$x_{t+1}^1$}}; \\
\node[] (intputs) {\large{Scond inputs}};    \& \node[] (xtm21) {\Large{$x_{t-1}^2$}};  \& \node[] (xt2) {\Large{$x_{t}^2$}}; \& \node[] (xtp21) {\Large{$x_{t+1}^2$}}; \\
% ...
\node[] () {\large{\dots}};  \& \node[] () {\Large{\dots}};  \& \node[] () {\Large{\dots}}; \& \node[] () {\Large{\dots}}; \\
\node[] (intputs) {\large{$N^{\text{th}}$ inputs}};    \& \node[] (xtmN1) {\Large{$x_{t-1}^N$}};  \& \node[] (xtN) {\Large{$x_{t}^N$}}; \& \node[] (xtpN1) {\Large{$x_{t+1}^N$}}; \\
};


%
\draw[-triangle 45, line width=0.35mm] (htm1N) -- node[midway,above] {} node[midway,below] {} (ytm1);
\draw[-triangle 45, line width=0.35mm] (htN) -- node[midway,above] {} node[midway,below] {} (yt);
\draw[-triangle 45, line width=0.35mm] (htp1N) -- node[midway,above] {} node[midway,below] {} (ytp1);
\draw[-triangle 45, line width=0.35mm] (xtm11) -- node[midway,above] {} node[midway,below] {} (htm11);
\draw[-triangle 45, line width=0.35mm] (xt1) -- node[midway,above] {} node[midway,below] {} (ht1);
\draw[-triangle 45, line width=0.35mm] (xtp11) -- node[midway,above] {} node[midway,below] {} (htp11);
\draw[-triangle 45, line width=0.35mm] (h1) -- node[midway,above] {} node[midway,below] {} (htm11);
\draw[-triangle 45, line width=0.35mm] (htm11) -- node[midway,above] {} node[midway,below] {} (ht1);
\draw[-triangle 45, line width=0.35mm] (ht1) -- node[midway,above] {} node[midway,below] {} (htp11);
\draw[-triangle 45, line width=0.35mm] (h2) -- node[midway,above] {} node[midway,below] {} (htm12);
\draw[-triangle 45, line width=0.35mm] (htm12) -- node[midway,above] {} node[midway,below] {} (ht2);
\draw[-triangle 45, line width=0.35mm] (ht2) -- node[midway,above] {} node[midway,below] {} (htp12);
\draw[-triangle 45, line width=0.35mm] (hN) -- node[midway,above] {} node[midway,below] {} (htm1N);
\draw[-triangle 45, line width=0.35mm] (htm1N) -- node[midway,above] {} node[midway,below] {} (htN);
\draw[-triangle 45, line width=0.35mm] (htN) -- node[midway,above] {} node[midway,below] {} (htp1N);
\draw[-triangle 45, line width=0.35mm]
    (xtm21) edge[bend left=30]  (htm12)
    (xt2) edge[bend left=30]  (ht2)
    (xtp21) edge[bend left=30]  (htp12)
    (xtmN1) edge[bend left=20]  (htm1N)
    (xtN) edge[bend left=20]  (htN)
    (xtpN1) edge[bend left=20]  (htp1N)
    (htm11) edge[bend right=30]  (ytm1)
    (ht1) edge[bend right=30]  (yt)
    (htp11) edge[bend right=30]  (ytp1)
    (htm12) edge[bend right=30]  (ytm1)
    (ht2) edge[bend right=30]  (yt)
    (htp12) edge[bend right=30]  (ytp1);

\end{tikzpicture}
}
 \caption{Parallel Long Short-Term (PLSTM) neural network.}
 \label{fig:plstm}
\end{center}
\end{figure}

\section{Parallel Long Short-Term Memory (PLSTM)}
\label{sec:pLstm}

The BRNN neural architecture uses the same sequence $\x$ as an input for both forward and backward directions, which is useful for information from a single stream. The paper proposes an original neural network, called Parallel RNN (PRNN) and presented in Figure~\ref{fig:plstm}, that takes advantage from the BRNN structure in a multistream context. By replacing the PRNN cells with LSTM cells, the proposed Parallel LSTM (PLSTM) is obtained.

The original PLSTM architecture corresponds to the PRNN description by replacing the $\mathcal{H}$ function with the LSTM composite function. PLSTM differs from the classical BLSTM by feeding forward, not a shared sequence, but different input vectors through a dedicated hidden layer $\h^n$ for each input vector $\x^n$. Moreover, BLSTM employs only 2 hidden layers due to its bidirectional concept while PLSTM can use multiple ones. The input sequences are considered independent and require to be mapped in homogeneous separate subspaces ($\W$ matrix from input $\x$ to hidden $\h$ spaces). Therefore, a single LSTM containing concatenated inputs from different independent sequences is not theoretically suitable for finding out a common homogeneous subspace to map heterogeneous input representation of parallel sequences.

Thus, for each $n^{th}$ stream ($1\le n \le N$), the PLSTM takes the input sequence $\x^n=(x_1^n,x_2^n,\dots,x_T^N)$ and computes the hidden sequence $\h^n=(h_1^n,h_2^n,\dots,h_T^N)$ and the output vector $\y$ by iterating from $t=1$ to $T$. 

\begin{align}
h_t^N&=\mathcal{H}(\W_{x^Nh^N}x_t^N+\W_{h^Nh^N}h^N_{t-1}+b_h^N) \\
&\dots\dots\dots\dots\dots\dots\dots\dots\dots\dots \\\
h_t^2&=\mathcal{H}(\W_{x^2h^2}x_t^2+\W_{h^2h^2}h^2_{t-1}+b_h^2) \\
h_t^1&=\mathcal{H}(\W_{x^1h^1}x_t^1+\W_{h^1h^1}h^1_{t-1}+b_h^1) \\
y_t&=\sum\limits_{n=1}^N  \W_{h^ny}h^n_t+b_y
\end{align}

where $N$ is the number of streams. In our experiments, the output vector $\y$ takes advantage of the $N$ channels to predict the telecast's genre for one given channel $n$ ($1\le n \le N$).
Therefore, PLSTM feeds forward separate sequences in order to predict a label and codes internal hidden structures between the parallel hidden sequences. 
\cite{graves2005framewise} introduces the BLSTM with Back Propagation Trough Time (BPTT) algorithm~\cite{schuster1999supervised} for training. For our proposed PLSTM architecture, the training takes place over $N$ input sequences:
\\{\bf Forward Pass}: feeds all input data for the sequences into the PLSTM and determines the predicted outputs.
\begin{itemize}
\item Do forward pass for the forward states of each of the $N$ layers.
\item Do forward pass for output layer.
\end{itemize} 
{\bf Backward Pass}: processes the error function derivative for the sequences used in the forward pass.
\begin{itemize}
\item Do backward pass for output neurons.
\item Do backward pass for forward states.
\end{itemize}
{\bf Updating Weights}
  
\section{Experimental Protocol}
\label{sec:experiments}

Multistream sequence classification is evaluated with the proposed PLSTM architecture (2 and 4 parallel sequences) as well as the classic LSTM network on an 
automatic TV show genre labeling task. Two n-gram based models (baseline) are also considered for fair comparison. Next sections describe the dataset, the genre sequence classification as well as the neural networks settings.

\subsection{Multichannel EPG dataset}
\label{sec:epgCorpus}

The Electronic Program Guide (EPG) dataset is extracted from 4 French TV channels (M6, TF1, France 5 and TV5 Monde) for 3 years, from January 2013 to December 2015. M6 channel is used in our experiments as the output stream.
Data from \textit{2013} and \textit{2014} are merged and split into the \textit{training} (70\%) and \textit{validation} (30\%) datasets using a \textit{stratified shuffle split}~\cite{pedregosa2011scikit} in order to preserve the same percentage of samples of each class in the output of both folds, while the \textit{2015} dataset is kept for testing. 
In order to guarantee a clean experimental environment, labels ({\it i.e.} genres) that are absent at least in one of the three folds were removed. 
Doing so allows us to have equivalent datasets in terms of labels vocabulary.
Table~\ref{tab:genresDistribution} shows the genres distribution for M6, the chosen output channel.
%

\begin{table}[!h]
\centering
\scalebox{0.8}{
\begin{tabular}{|c||c|c|c|}
\hline
\textbf{Genres} & \textbf{Training} & \textbf{Validation} & \textbf{Test} \\ \hline
Weather & 2,691 & 1,153 & 1,683 \\ \hline
Fiction & 1,890 & 810 & 1,444 \\ \hline
News & 913 & 392 & 663 \\ \hline 
Other magazine & 981 & 421 & 451 \\ \hline
Music & 461 & 197 & 330 \\ \hline
Teleshopping & 421 & 180 & 307 \\ \hline
TV game show & 476 & 204 & 284 \\ \hline
Cartoon & 361 & 155 & 205 \\ \hline
Other & 277 & 119 & 129 \\ \hline
Reality TV & 83 & 36 & 76 \\ \hline
Documentary & 29 & 13 &14 \\ \hline \hline
{\bf Total} & 8,583 & 3,680 & 5,586 \\ \hline
\end{tabular}
}
\caption{Genres Distribution for train, validation and test sets in M6 channel output.}
\label{tab:genresDistribution}
\end{table}

\subsection{Genre Prediction Experiments}
\label{sec:genrePredictionExperiments}
For a given input history sequence (composed of the $n$ previous telecast genres), a genre label representing the next M6's telecast is output. The size of the genre sequences ($n$) varies from $1$ to $4$. Then, three input configurations are employed. {\bf Mono-channel input:} only M6 history sequences for a baseline \textit{n-gram} experiment (with a \textit{statistical language model} from the SRILM toolkit~\cite{stolcke2002srilm}) and a straightforward \textit{LSTM} model. {\bf Bi-channel input:} both M6 and TF1 channel histories are employed as input for \textit{P2LSTM} (PLSTM with two parallel streams as a BLSTM with forward-forward directions and separate inputs). The aim of this experiment is to move up the context's information using a similar and rival channel, the two being generalist channels. {\bf Multichannel input:} History of each of the 4 streams ({\it i.e.} channels) is used as input for \textit{4n-gram} and P4LSTM experiments (PLSTM with 4 parallel streams).

\subsection{Neural Networks Setup}
\label{sec:neuralNetworksSetup}

The classical LSTM, and the proposed P2LSTM and P4LSTM, are composed with 3 layers: input layer $\x$ of size varying from $1$ to $4$, a hidden layer $\h$ of size $80$ for all LSTM-based models and an output layer $\y$ with a size equals to the number of different possible TV genres ($11$). The Keras library~\cite{kerasChollet2015}, based on Theano~\cite{Bastien-Theano-2012} for fast tensor manipulation and CUDA-based GPU acceleration, has been employed to train neural networks on an Nvidia GeForce GTX TITAN X GPU card. The training times, detailed in Table~\ref{tab:runningDuration} for all models, match with the sequence size of all models. Indeed, even with the most time-consuming configuration, namely P4LSTM with $4$ elements history, the training does not last more than $25$ minutes.
%

\begin{table}[!ht]
\centering
\scalebox{0.8}{
\begin{tabular}{|l||c|c|c|c|c|}
\hline
\textbf{Sequence size} & \textbf{1} & \textbf{2} & \textbf{3} & \textbf{4} \\ \hline
\textbf{n-gram} & 1 & 1 & 1 & 1 \\ \hline
\textbf{4n-gram} & 2 & 5 & 17 & 51 \\ \hline
\textbf{LSTM} & 51 & 146 & 319 & 362 \\ \hline
\textbf{P2LSTM} & 259 & 473 & 485 & 439 \\ \hline
\textbf{P4LSTM} & 536 & 923 & 844 & 1,386 \\ \hline
\end{tabular}
}
\caption{Training times (in seconds) of models employed during the experiments for different telecast genres sequence sizes.}
\label{tab:runningDuration}
\end{table}


\section{Results and Discussion}
\label{s:Results}

Table~\ref{tab:f1} shows the overall results, in terms of the standard F1 metric related to the genre prediction task outputs, using each method and for different stream sequence sizes from $1$ to $4$. 

\begin{table}[!htbp]
\centering
\scalebox{0.8}{
\begin{tabular}{|c||c|c||c|c|c|}
\hline
\textbf{Seq. size} & \textbf{n-gram} & \textbf{4n-gram} & \textbf{LSTM} & \textbf{P2LSTM} & \textbf{P4LSTM} \\ \hline
1 & 18.97 & 59.60 & 11.46 & 47.57 & 45.66 \\ \hline
2 & 51.25 & 58.36 & 46.49 & 55.09 & 62.76 \\ \hline
3 & 57.34 & 57.16 & 55.64 & 58.68 & 59.80 \\ \hline
4 & 55.89 & 57.34 & 58.15 & 60.77 & \textbf{66.04} \\ \hline
\end{tabular}
}
\caption{F1-score ($\%$) of each n-gram and LSTM models.}
\label{tab:f1}
\end{table}

\subsection{N-gram based models}
\label{sec:ngramBasedExperiments}

The multi-channel 4n-gram model outperforms the simple n-gram one for each of the different 4 genre sequence configurations except for $3$ sized history. \textit{4n-gram} method reaches around $60\%$ of F-score using $1$ sized sequences against near $57\%$ for mono-channel n-gram using its best history configuration. The observed results confirm the interest of using multiple streams to predict the next telecast's genre for a specified channel.

\subsection{LSTM and PLSTM}
\label{sec:lstmAndPlstm}
One can figure out from Table~\ref{tab:f1} that mono-channel LSTM performances gradually become closer and closer to the multichannel n-gram model ones when the size of sequences moves up and overtakes it with an F1 score of $58\%$ using $4$ sized sequences. Therefore, LSTM-based models require longer sequences to learn long term dependencies than the n-gram based methods. P4LSTM obtains the best result with an F1 score close of $66\%$ using a sequence of size $4$. In order to analyze these results, the Error Rates (ER) are also presented in Table~\ref{tab:errorRate}. The overall F1 scores are different from those related to the ER. For example, at its best configuration of a $4$ sized sequence, P4LSTM error rate reaches about $21.5\%$, which corresponds to a correct rate of $78.5\%$ against an F1-measure of only $66\%$. The reason of this is that the F1-metric may be not suitable for the task due to the labels imbalance with different numbers of genre occurrences varying from 14 to 1,683 in the test set.

\begin{table}[!htbp]
\centering
\scalebox{0.8}{
\begin{tabular}{|c||c|c||c|c|c|}
\hline
\textbf{Seq. size} & \textbf{n-gram} & \textbf{4n-gram} & \textbf{LSTM} & \textbf{P2LSTM} & \textbf{P4LSTM} \\ \hline
1 & 51.52 & 30.08 & 63.03 & 36.47 & 35.25 \\ \hline
2 & 39.19 & 29.11 & 44.90 & 28.97 & 25.65 \\ \hline
3 & 31.60 & 29.72 & 31.69 & 27.60 & 24.01 \\ \hline
4 & 36.32 & 30.59 & 28.28 & 25.98 & \textbf{21.45} \\ \hline
\end{tabular}
}
\caption{Error rates (ER) observed for each n-gram and LSTM models for different sequence sizes.}
\label{tab:errorRate}
\end{table}

\subsection{Discussion}

Confusion matrices of 4n-gram and P4LSTM experiments using sequences of $4$ telecasts are shown in Tables~\ref{tab:confMat4ngram} and~\ref{tab:confMatP4lstm} to point out  benefits of the proposed PLSTM model.


\begin{table}[!h]
\centering
\scalebox{0.71}{
\begin{tabular}{|l|c|c|c|c|c|c|c|c|c|c|c|}
\hline
\textbf{Weather} & 1257 & {\color[HTML]{009901}65} & 0 & {\color[HTML]{009901}320} & 11 & 4 & 17 & 0 & 4 & 2 & 3 \\ \hline
\textbf{Fiction} & {\color[HTML]{009901} 291} & 875 & 0 & {\color[HTML]{009901}175} & 35 & 9 & 15 & 26 & 14 & 0 & 4 \\ \hline
\textbf{News} & 2 & 12 & 623 & 1 & 24 & 0 & 0 & 0 & 1 & 0 & 0 \\ \hline
\textbf{Other mag.} & {\color[HTML]{009901}66} & {\color[HTML]{009901}38} & 3 & \textbf{289} & 8 & 2 & 6 & 0 & 28 & 6 & 5 \\ \hline
\textbf{Music} & {\color[HTML]{009901}16} & {\color[HTML]{009901}34} & 17 & 5 & 215 & 0 & 3 & 2 & 38 & 0 & 0 \\ \hline
\textbf{Teleshop.} & {\color[HTML]{009901}45} & 1 & 0 & 1 & 0 & 245 & 0 & 0 & 15 & 0 & 0 \\ \hline
\textbf{TV game sh.} & 2 & 6 & 0 & 19 & 0 & 0 & \textbf{243} & 0 & 6 & 8 & 0 \\ \hline
\textbf{Cartoon} & 0 & 32 & 6 & 0 & 9 & \textbf{\textit{\underline{102}}} & 0 & 56 & 0 & 0 & 0 \\ \hline
\textbf{Other} & 9 & 10 & 0 & {\color[HTML]{009901}24} & 1 & 2 & 15 & 0 & \textbf{64} & 2 & 2 \\ \hline
\textbf{Reality TV} & {\color[HTML]{009901}20} & 14 & 0 & 9 & 0 & 0 & 4 & 0 & 19 & \textbf{9} & 1 \\ \hline
\textbf{Docum.} & {\color[HTML]{009901}4} & 3 & 0 & {\color[HTML]{009901}4} & 0 & 0 & 0 & 0 & 2 & 0 & \textbf{1} \\ \hline
\end{tabular}
}
\caption{Confusion matrix for the 4n-gram output using $4$ sized sequences: labels are shown according to their decreasing frequency as in Table~\ref{tab:genresDistribution}.}
\label{tab:confMat4ngram}
\end{table}

\begin{table}[!h]
\centering
\scalebox{0.7}{
\begin{tabular}{|l|c|c|c|c|c|c|c|c|c|c|c|}
\hline
\textbf{Weather} & \textbf{1600} & {\color[HTML]{009901}52} & 0 & {\color[HTML]{009901}24} & 0 & 6 & 0 & 0 & 1 & 0 & 0 \\ \hline
\textbf{Fiction} & {\color[HTML]{009901}343} & \textbf{937} & 0 & {\color[HTML]{009901}109} & 19 & 0 & 21 & 11 & 4 & 0 & 0 \\ \hline
\textbf{News} & 6 & 1 & \textbf{652} & 0 & 3 & 1 & 0 & 0 & 0 & 0 & 0 \\ \hline
\textbf{Other mag.} & {\color[HTML]{009901}92} & {\color[HTML]{009901}51} & 0 & 269 & 5 & 0 & 7 & 0 & 27 & 0 & 0 \\ \hline
\textbf{Music} & 13 & 20 & 2 & 1 & \textbf{290} & 0 & 3 & 0 & 1 & 0 & 0 \\ \hline
\textbf{Teleshop.} & {\color[HTML]{009901}61} & 0 & 0 & 0 & 0 & \textbf{246} & 0 & 0 & 0 & 0 & 0 \\ \hline
\textbf{TV game sh.} & {\color[HTML]{009901}39} & 9 & 0 & 6 & 17 & 0 & 196 & 0 & 17 & 0 & 0 \\ \hline
\textbf{Cartoon} & 2 & {\color[HTML]{009901}40} & 0 & 0 & 0 & 5 & 0 & \textbf{158} & 0 & 0 & 0 \\ \hline
\textbf{Other} & {\color[HTML]{009901}42} & 10 & 0 & {\color[HTML]{009901}31} & 0 & 1 & 6 & 0 & 39 & 0 & 0 \\ \hline
\textbf{Reality TV} & {\color[HTML]{009901}16} & 3 & 0 & {\color[HTML]{009901}23} & 0 & 0 & 13 & 0 & 20 & 1 & 0 \\ \hline
\textbf{Docum.} & {\color[HTML]{009901}4} & 2 & 0 & {\color[HTML]{009901}5} & 0 & 0 & 1 & 0 & 2 & 0 & 0 \\ \hline
\end{tabular}
}
\caption{Confusion matrix for the P4LSTM output using $4$ sized sequences: labels are shown according to their decreasing frequency as in Table~\ref{tab:genresDistribution}.}
\label{tab:confMatP4lstm}
\end{table}

It is worth emphasizing that most of the missed instances in all systems are wrongly labeled as one of the two most frequent classes, \textit{Weather} and \textit{Fiction}, as well as the \textit{Other Magazine} genre (some examples are in green cells). 
\textit{False positives} are more recurrent in \textit{Other Magazine} than in \textit{News}, the relatively more frequent class. The reason is that \textit{News} is a well defined genre occurring mostly at the same time each day unlike \textit{Other Magazine} genre that encompasses various telecasts that are broadcast at several and irregular daytime. 
\textit{Teleshopping} shows are often broadcast at nearly the same time of morning than \textit{Cartoons} and affects dramatically the performance of the 4n-gram model in this context (cf. underlined italic cell in Table~\ref{tab:confMat4ngram}). 
Finally, the confusion matrix of P4LSTM experiment shows that this system fails more dramatically to predict the least frequent genres \textit{Others}, \textit{Reality TV}, and \textit{Documentary}. For example, for the two least frequent genres, \textit{Reality TV} and \textit{Documentary}, respectively no more than one of the $76$ and the $14$ instances was correctly found. This leads to a \textit{precision} and a \textit{recall} of $0$ which penalizes their averages respectively and then the overall F-score.

\begin{table}[!htbp]
\centering
\scalebox{0.8}{
\begin{tabular}{|c||c|c||c|c|c|}
\hline
\textbf{Seq. size} & \textbf{n-gram} & \textbf{4n-gram} & \textbf{LSTM} & \textbf{P2LSTM} & \textbf{P4LSTM} \\ \hline
1 & 23.18 & 70.58 & 14.00 & 58.15 & 55.80 \\ \hline
2 & 59.31 & 61.64 & 56.83 & 67.32 & 71.80 \\ \hline
3 & 66.63 & 66.61 & 68.00 & 71.72 & 73.09 \\ \hline
4 & 64.48 & 66.86 & 71.07 & 74.27 & \textbf{75.75} \\ \hline
\end{tabular}
}
\caption{F1 score ($\%$) of n-gram and LSTM models, the two least frequent genres \textit{Reality TV} and \textit{Documentary} not being included.}
\label{tab:f1_-2LFG}
\end{table}


\begin{figure}[!ht]
\begin{center}
\scalebox{0.9}{
\begin{tikzpicture}
\begin{axis}[
    width=9cm, height=6cm,   
    grid = major,
    grid style={dashed, gray!30},
    xmin=1,   
    xmax=4,  
    ymin=9,   
    ymax=80,  
    axis background/.style={fill=white},
    xtick={1,2,3,4},
    xlabel=Sequence size,
    xticklabels={\textcolor{red}{1},2,3,\textcolor{greenForest}{4}},
    tick align=outside,
    mark repeat={600},
 legend style={draw=none},
	legend columns=3, 
        legend style={/tikz/column 2/.style={ column sep=1pt,},},
        legend style={at={(1.0,0.11)},anchor=east,font=\small}]
\draw[very thick,greenForest] (235,410) node (A) {{\large Max = 75.75}}; 
\draw[very thick,greenForest] (300,666) circle (0.10cm);
\draw[very thick,greenForest] (300,666) node (B) {};
\draw[->,very thick,greenForest] (A)--(B);
\addplot+[mark=none,color=black] coordinates {
 (1, 23.18)
 (2, 59.31)
 (3, 66.63)
 (4, 64.48)};\addlegendentry{n-gram}
%
\addplot+[mark=none,color=black, dashed] coordinates {
 (1, 70.58)
 (2, 61.64)
 (3, 66.61)
 (4, 66.86)};\addlegendentry{4n-gram}
\addplot+[mark=none,color=red] coordinates {
 (1, 14.00)
 (2, 56.83)
 (3, 68.00)
 (4, 71.07)};\addlegendentry{LSTM}
\addplot+[mark=none,color=purple] coordinates {
 (1, 58.15)
 (2, 67.32)
 (3, 71.72)
 (4, 74.27)};\addlegendentry{P2LSTM}
\addplot+[mark=none,color=greenForest] coordinates {
 (1, 55.80)
 (2, 71.80)
 (3, 73.09)
 (4, 75.75)};\addlegendentry{P4LSTM}
\end{axis}
\end{tikzpicture}
}
\caption{F1 score for n-gram and LSTM models, the two least frequent genres \textit{Reality TV} and \textit{Documentary} not being included.}
\label{fig:4n-gram_vs_lstm}
\end{center}
\end{figure}
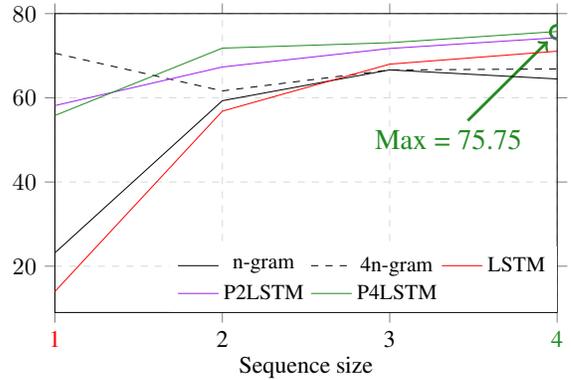


In order to evaluate the impact of the least frequent genres on the F1 measure, especially on the PLSTM systems, we also reported on Table~\ref{tab:f1_-2LFG} the F1 results on the same outputs of the experiments of Table~\ref{tab:f1} by excluding the two least frequent genres from the averages of precision and recall (\textit{Reality TV} and \textit{Documentary}).

Overall, the results of the PLSTM detailed in Table~\ref{tab:f1_-2LFG} and Figure~\ref{fig:4n-gram_vs_lstm}, demonstrate the benefits obtained at least for history sequences longer than $2$ genres with an F1 score greater than $71\%$.

Regarding multichannel P4LSTM approach, the highest performance reaches an F1-measure of about $76\%$ using $4$ sized sequences with a gain of about $2$ and $5$ points compared respectively to P2LSTM and 4n-gram model best performances. 




  
\section{Conclusion}
\label{sec:Conclusion}

The paper proposes an original Long Short-Term Memory (LSTM) based neural network architecture for automatic classification of multistream sequences called PLSTM. PLSTM is evaluated during a telecast genre prediction task and the observed results show that the proposed PLSTM is efficient when the size of sequences is large enough with a gain of more than $10$ points of error rate compared to classical n-gram model, and about $7$ and $4$ points respectively compared to LSTM and P2LSTM. Future works will apply this promising multistream neural network architecture to Spoken Language Understanding tasks such as topic extraction, keyword spotting and Part-of-Speech tagging. 

  \bibliographystyle{IEEEtran}

  \bibliography{slt_2016}

\begin{thebibliography}{10}
\providecommand{\url}[1]{#1}
\csname url@samestyle\endcsname
\providecommand{\newblock}{\relax}
\providecommand{\bibinfo}[2]{#2}
\providecommand{\BIBentrySTDinterwordspacing}{\spaceskip=0pt\relax}
\providecommand{\BIBentryALTinterwordstretchfactor}{4}
\providecommand{\BIBentryALTinterwordspacing}{\spaceskip=\fontdimen2\font plus
\BIBentryALTinterwordstretchfactor\fontdimen3\font minus
  \fontdimen4\font\relax}
\providecommand{\BIBforeignlanguage}[2]{{%
\expandafter\ifx\csname l@#1\endcsname\relax
\typeout{** WARNING: IEEEtran.bst: No hyphenation pattern has been}%
\typeout{** loaded for the language `#1'. Using the pattern for}%
\typeout{** the default language instead.}%
\else
\language=\csname l@#1\endcsname
\fi
#2}}
\providecommand{\BIBdecl}{\relax}
\BIBdecl

\bibitem{gers2001applying}
F.~A. Gers, D.~Eck, and J.~Schmidhuber, ``Applying lstm to time series
  predictable through time-window approaches,'' in \emph{Artificial Neural
  Networks—ICANN 2001}.\hskip 1em plus 0.5em minus 0.4em\relax Springer,
  2001, pp. 669--676.

\bibitem{severyn2015twitter}
A.~Severyn and A.~Moschitti, ``Twitter sentiment analysis with deep
  convolutional neural networks,'' in \emph{Proceedings of the 38th
  International ACM SIGIR Conference on Research and Development in Information
  Retrieval}.\hskip 1em plus 0.5em minus 0.4em\relax ACM, 2015, pp. 959--962.

\bibitem{huang2016modeling}
M.~Huang, Y.~Cao, and C.~Dong, ``Modeling rich contexts for sentiment
  classification with lstm,'' \emph{CoRR}, vol. abs/1605.01478, 2016.

\bibitem{lecun1998gradient}
Y.~LeCun, L.~Bottou, Y.~Bengio, and P.~Haffner, ``Gradient-based learning
  applied to document recognition,'' \emph{Proceedings of the IEEE}, vol.~86,
  no.~11, pp. 2278--2324, 1998.

\bibitem{elman1990finding}
J.~L. Elman, ``Finding structure in time,'' \emph{Cognitive science}, vol.~14,
  no.~2, pp. 179--211, 1990.

\bibitem{hochreiter1997long}
S.~Hochreiter and J.~Schmidhuber, ``Long short-term memory,'' \emph{Neural
  computation}, vol.~9, no.~8, pp. 1735--1780, 1997.

\bibitem{graves2005framewise}
A.~Graves and J.~Schmidhuber, ``Framewise phoneme classification with
  bidirectional lstm and other neural network architectures,'' \emph{Neural
  Networks}, vol.~18, no.~5, pp. 602--610, 2005.

\bibitem{sundermeyer2012lstm}
M.~Sundermeyer, R.~Schl{\"u}ter, and H.~Ney, ``Lstm neural networks for
  language modeling.'' in \emph{INTERSPEECH}, 2012, pp. 194--197.

\bibitem{vinyals2015show}
O.~Vinyals, A.~Toshev, S.~Bengio, and D.~Erhan, ``Show and tell: A neural image
  caption generator,'' in \emph{Proceedings of the IEEE Conference on Computer
  Vision and Pattern Recognition}, 2015, pp. 3156--3164.

\bibitem{graves2005bidirectional}
A.~Graves, S.~Fern{\'a}ndez, and J.~Schmidhuber, ``Bidirectional lstm networks
  for improved phoneme classification and recognition,'' in \emph{Artificial
  Neural Networks: Formal Models and Their Applications--ICANN 2005}.\hskip 1em
  plus 0.5em minus 0.4em\relax Springer, 2005, pp. 799--804.

\bibitem{fernandez2007application}
S.~Fern{\'a}ndez, A.~Graves, and J.~Schmidhuber, ``An application of recurrent
  neural networks to discriminative keyword spotting,'' in \emph{Artificial
  Neural Networks--ICANN 2007}.\hskip 1em plus 0.5em minus 0.4em\relax
  Springer, 2007, pp. 220--229.

\bibitem{wollmer2008abandoning}
M.~W{\"o}llmer, F.~Eyben, S.~Reiter, B.~Schuller, C.~Cox, E.~Douglas-Cowie, and
  R.~Cowie, ``Abandoning emotion classes-towards continuous emotion recognition
  with modelling of long-range dependencies.'' in \emph{INTERSPEECH}, vol.
  2008.\hskip 1em plus 0.5em minus 0.4em\relax Citeseer, 2008, pp. 597--600.

\bibitem{bourlard2012connectionist}
H.~A. Bourlard and N.~Morgan, \emph{Connectionist speech recognition: a hybrid
  approach}.\hskip 1em plus 0.5em minus 0.4em\relax Springer Science \&
  Business Media, 2012, vol. 247.

\bibitem{bengio1999markovian}
Y.~Bengio, ``Markovian models for sequential data,'' \emph{Neural computing
  surveys}, vol.~2, no. 1049, pp. 129--162, 1999.

\bibitem{gers2003learning}
F.~A. Gers, N.~N. Schraudolph, and J.~Schmidhuber, ``Learning precise timing
  with lstm recurrent networks,'' \emph{The Journal of Machine Learning
  Research}, vol.~3, pp. 115--143, 2003.

\bibitem{schuster1997bidirectional}
M.~Schuster and K.~K. Paliwal, ``Bidirectional recurrent neural networks,''
  \emph{Signal Processing, IEEE Transactions on}, vol.~45, no.~11, pp.
  2673--2681, 1997.

\bibitem{schuster1999supervised}
M.~Schuster, ``On supervised learning from sequential data with applications
  for speech recognition,'' \emph{Daktaro disertacija, Nara Institute of
  Science and Technology}, 1999.

\bibitem{pedregosa2011scikit}
F.~Pedregosa, G.~Varoquaux, A.~Gramfort, V.~Michel, B.~Thirion, O.~Grisel,
  M.~Blondel, P.~Prettenhofer, R.~Weiss, V.~Dubourg \emph{et~al.},
  ``Scikit-learn: Machine learning in python,'' \emph{The Journal of Machine
  Learning Research}, vol.~12, pp. 2825--2830, 2011.

\bibitem{stolcke2002srilm}
A.~Stolcke \emph{et~al.}, ``Srilm-an extensible language modeling toolkit.'' in
  \emph{INTERSPEECH}, vol. 2002, 2002, p. 2002.

\bibitem{kerasChollet2015}
F.~Chollet, ``keras,'' \url{https://github.com/fchollet/keras}, 2015.

\bibitem{Bastien-Theano-2012}
F.~Bastien, P.~Lamblin, R.~Pascanu, J.~Bergstra, I.~J. Goodfellow, A.~Bergeron,
  N.~Bouchard, and Y.~Bengio, ``Theano: new features and speed improvements,''
  Deep Learning and Unsupervised Feature Learning NIPS 2012 Workshop, 2012.

\end{thebibliography}


\end{document}